\documentclass[conference]{IEEEtran}

\IEEEoverridecommandlockouts
\usepackage{multirow}
\newcommand{\RNum}[1]{\uppercase\expandafter{\romannumeral #1\relax}}
\usepackage{cite}
\usepackage{arydshln}
\usepackage{makecell}
\usepackage[table]{xcolor}
\definecolor{mycolor}{RGB}{30,220,220}
\usepackage{float}
\usepackage[caption = false]{subfig}
\usepackage[final]{graphicx}
\usepackage{amsmath,amssymb,amsfonts}
\usepackage{algorithm2e}
\usepackage{mathtools}
\usepackage{textgreek}
\usepackage{relsize}
\usepackage{graphicx}
\usepackage{textcomp}
\usepackage{arydshln}
\usepackage[utf8]{inputenc}
\usepackage{amsmath,amssymb,amsthm}
\usepackage{pifont}
\usepackage[a4paper, total={184mm,239mm}]{geometry}
\def\BibTeX{{\rm B\kern-.05em{\sc i\kern-.025em b}\kern-.08em
    T\kern-.1667em\lower.7ex\hbox{E}\kern-.125emX}}

\begin{document}
\title{DNN Memory Footprint Reduction via Post-Training Intra-Layer Multi-Precision Quantization}
% \author{}
\author{\IEEEauthorblockN{Behnam Ghavami\textsuperscript{1,2}, Amin Kamjoo, Lesley Shannon\textsuperscript{2}, Steve Wilton\textsuperscript{1}}
\IEEEauthorblockA{\textsuperscript{1}University of British Columbia, Canada\\ \textsuperscript{2}Simon~Fraser~University, Canada\\
\\
"This paper has been accepted for presentation at the 25th International Symposium on Quality Electronic Design (ISQED'24)."
}
}
% \author{\IEEEauthorblockN{1\textsuperscript{st} Behnam Ghavami}
% \and
% \IEEEauthorblockN{2\textsuperscript{nd} Amin Kamjoo}
% Kerman, Iran \\
% amin.kamjooo@gmail.com}
% \and
% \IEEEauthorblockN{3\textsuperscript{rd} Lesley Shannon}
% \and
% \IEEEauthorblockN{4\textsuperscript{th} Steve Wilton}
% email address or ORC ID}
% }
%Non-overlapping Regions and Different Number of Bits} Deep Neural Networks
% \thanks{}
\maketitle

% Compacting DNNs using Post-Training Mixed-Precision, Multi-Scheme Piecewise Linear Quantization 
% \author{\IEEEauthorblockN{1\textsuperscript{st} Given Name Surname}}
% \and
% \and
% \and
% \and
% \and

% \maketitle
\vspace{-0.05in}
\begin{abstract}
The imperative to deploy Deep Neural Network (DNN) models on resource-constrained edge devices, spurred by privacy concerns, has become increasingly apparent. To facilitate the transition from cloud to edge computing, this paper introduces a technique that effectively reduces the memory footprint of DNNs, accommodating the limitations of resource-constrained edge devices while preserving model accuracy. Our proposed technique, named Post-Training Intra-Layer Multi-Precision Quantization (PTILMPQ), employs a post-training quantization approach, eliminating the need for extensive training data. By estimating the importance of layers and channels within the network, the proposed method enables precise bit allocation throughout the quantization process. Experimental results demonstrate that PTILMPQ offers a promising solution for deploying DNNs on edge devices with restricted memory resources. For instance, in the case of ResNet50, it achieves an accuracy of 74.57\% with a memory footprint of 9.5 MB, representing a 25.49\% reduction compared to previous similar methods, with only a minor 1.08\% decrease in accuracy.
\end{abstract}

% to address memory constraints when deploying DNNs on resource-constrained edge devices.

% \begin{abstract}
% This document is a model and instructions for \LaTeX.
% This and the IEEEtran.cls file define the components of your paper [title, text, heads, etc.]. *CRITICAL: Do Not Use Symbols, Special Characters, Footnotes, 
% or Math in Paper Title or Abstract.
% \end{abstract}
\vspace{0.1in}

%%%%%%%%%%%%%%

% \begin{document}

% \title{\ToolName: Error Resilience Design of Deep Neural Networks via a  Fine-grained Trainable Activation Function}

% \input{abstract}
% \maketitle
%

% \input{abstract}
\section{Introduction}\label{sec:intro}
In recent years, the prominence of Deep Neural Networks (DNNs) has surged, particularly in domains like machine vision, natural language processing, and speech recognition. These networks, characterized by their intricate architectures and numerous parameters, have demonstrated remarkable capabilities. However, their increasing complexity comes with a significant demand for computational resources and memory space. The challenge intensifies when deploying DNNs on devices with constrained resources, such as Internet of Things (IoT) devices \cite{9985008}. These compact devices not only grapple with limited memory but also face constraints in terms of energy consumption and processing power. Notably, the absence of external memory in many IoT devices necessitates that all DNN parameters be accommodated within the confines of Static Random-Access Memory (SRAM).
This convergence of intricate DNN architectures with the constraints of memory, energy, and processing in IoT devices underscores the pressing need for innovative solutions.

 % Efficient memory management strategies, lightweight model architectures, and optimized algorithms are pivotal in addressing these challenges, enabling the seamless integration of powerful DNNs into resource-constrained environments.

To tackle the challenges posed by memory constraints in edge devices, one strategic approach involves the design of more streamlined DNN architectures, exemplified by models such as MobileNet \cite{https://doi.org/10.48550/arxiv.1704.04861} and SqueezeNet \cite{iandola2016squeezenet}. While these architectures prove to be efficient, training them from scratch often incurs substantial computational costs. An alternative avenue is the utilization of compression techniques aimed at reducing the number of parameters in DNNs, encompassing both weights and activity functions \cite{cheng2017survey}. Compression methods fall into various categories, and one notable classification includes pruning methods \cite{DBLP:journals/corr/LiKDSG16, theis2018faster}, knowledge distillation \cite{hinton2015distillingnew, yim2017gift}, low-rank factorization \cite{lowrank, povey2018semi}, and quantization \cite{zhou2017balanced, krishnamoorthi2018quantizing, nagel2019data}.
% These compression methods present a cost-effective alternative to redesigning DNN architectures. 

Quantization is a favored compression method as it reduces memory requirements without necessitating significant changes to the model architecture. However, a common drawback quantization is the potential diminishment of DNN accuracy due to the loss of parameters. 
% This trade-off between quantization efficiency and model accuracy underscores the need for a careful balance and underscores the ongoing exploration of innovative methodologies to mitigate such trade-offs.  
To address the accuracy loss following quantization, several methods have been proposed to compensate for this drawback. One such approach involves retraining the DNN using the original training dataset, a process known as quantization-aware training (QAT) \cite{dong2019hawq}. Retraining facilitates the recovery of some of the lost accuracy, and while QAT can achieve higher accuracy compared to other methods, it comes with the overhead of requiring a training dataset and retraining processes. In contrast, the post-training quantization (PTQ) technique is employed to prevent the decrease in DNN accuracy without necessitating additional training steps \cite{krishnamoorthi2018quantizing}. While QAT may achieve superior accuracy, PTQ offers practical benefits that make it a favorable choice in scenarios where privacy, security, and cost considerations take precedence over marginal gains in model performance. 

Within the domain of PTQ, various methods have been proposed to address the challenge of accuracy loss. Cai et al.  \cite{cai2020zeroq} utilized a weight simulation approach, generating synthetic data resembling real-world data to facilitate the retraining of the DNN and recover lost accuracy; however, this method introduces overhead due to the creation of artificial data and the associated retraining process. Fang et al. \cite{fang2020post} employed non-overlapping regions during quantization to preserve accuracy by categorizing weights within specific intervals based on their values, achieving precise quantization within distinct areas. Nevertheless, a limitation of this approach is its use of uniform single-precision quantization for all weights, neglecting the potential advantages of mixed precision, particularly valuable in scenarios with memory constraints.

% These recent developments underscore the dynamic landscape of PTQ techniques, each presenting unique trade-offs and optimizations. As the field continues to evolve, finding a delicate balance between accuracy preservation, computational efficiency, and memory utilization remains a key focus in enhancing the adaptability of compressed DNNs for deployment on resource-constrained IoT devices.

% In the realm of the PTQ technique, methods have been devised to address accuracy loss. "ZEROQ" \cite{cai2020zeroq} utilizes weight simulation by generating synthetic data similar to real-world data. This synthetic data aids in retraining the DNN to compensate for lost accuracy without compromising security or privacy, as it doesn't require access to the training dataset. However, this technique incurs overhead due to artificial data creation and the retraining process.
% Recently, "PWLQ" \cite{fang2020post} employs non-overlapping regions to preserve accuracy. By identifying breakpoints in the bell-shaped weight distribution, these regions are strategically determined based on quantization error. Weights are placed within specific intervals according to their values, enabling precise quantization within distinct areas. However, this method employs uniform single-precision quantization for all weights, missing out on the benefits of mixed precision, which can be crucial in memory-constrained scenarios.

This paper addresses the aforementioned limitation by introducing a novel approach that integrates mixed precision and meticulously evaluates the significance of each layer. The proposed method, named Post-Training Intra-Layer Multi-Precision Quantization (PTILMPQ), aims to synergize the benefits of both precision and layer importance to effectively mitigate the accuracy loss associated with quantization, particularly focusing on non-overlapping regions. We categorize layers into two classes: important and non-important, based on the values of their weight tensors. Similarly, within each layer, channels are segmented into important and non-important categories. The distinction lies in our strategy: for important layers, we opt for high-precision channels, while non-important layers predominantly employ low or ultra-low bit precision channels. Beyond layer and channel differentiation, we introduce a further refinement by partitioning the weight distribution range within each layer into two distinct regions. The quantization process, consequently, takes into account two critical criteria: (1) the significance of the layer and channel in terms of precision, and (2) the positioning of the weight tensor within either the dense or sparse region. PTILMPQ aims to strike a balance that enhances DNN accuracy post-quantization, presenting a nuanced solution for practical deployment in scenarios with varied constraints and requirements.  To validate the effectiveness of the proposed method, we conducted experiments using 8-bit mixed precision for ResNet50 neural network (NN) weights and 8-bit precision for activation functions. The results demonstrate a substantial 37.74\% reduction in model size compared to the similar methods, which relies solely on non-overlapping regions. 

% Furthermore, when employing 4-bit mixed precision for ResNet50 NN weights and 8-bit precision for activation functions, our method achieves a 9.72\% reduction in model size compared to the methods, which exclusively utilizes mixed precision. Extending our evaluation to the MobileNetV2 model, where we apply 8-bit mixed precision for weights and 8-bit quantization for activation functions, our method showcases a commendable 24.84\% reduction in model size compared to the PWLQ approach.

% Our contributions in this paper encompass the following key advancements:
% \begin{itemize}
%     \item  We employ a method for identifying and preserving the most crucial layers and channels within the neural network, thereby reducing its overall size.
%     \item We ensure the preservation of neural network accuracy during the quantization process by utilizing non-overlapping regions for quantization operations.
% \end{itemize}

% This paper is organized as follows:
% First, the previous works are reviewed, then the proposed method is described. Finally, the proposed method is examined and compared with other methods of quantization.

The rest of this paper is as follows: Section II presents the proposed quantization method. Sections III, discuss the experimental setup and results. Finally, Section IV concludes the paper.

\section{Proposed Quantization Methodology}
\label{sec:method}
\begin{figure*}[h]
 	\begin{center}
 	    \vspace{-0.05in}
 		\includegraphics[width = 0.8\textwidth]{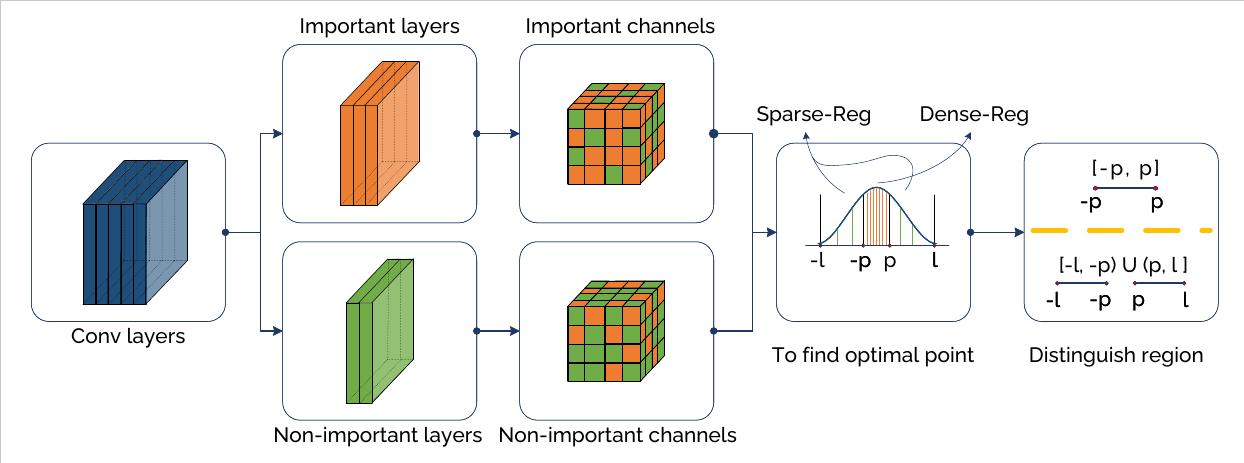}
 		\vspace{-0.1in}
 		\caption{ A general view of proposed method. Based on the function  $F_l$ (Equation \ref{eq:layers}), layers are divided into important and non-important categories. Selection of the most important channels is performed for each layer. The quantization process is performed according to important channels on non-overlapping regions (Sparse-Region and Dense-Region)}
 		\vspace{-0.1in}
 		\label{fig:all_phases}
 	\end{center}	
\end{figure*}

This section introduces our proposed quantization method, which comprises two fundamental phases (Figure \ref{fig:all_phases}):
\begin{itemize}
\item Layer and Channel Separation (Phase 1): In this initial phase, network layers are classified into two distinct groups: important and non-important. Similarly, channels within these layers are assigned corresponding categories. This classification informs the determination of the precise number of quantization bits required for each layer and channel.
\item Quantization using Non-Overlapping Regions (Phase 2): In the subsequent phase, non-overlapping regions are employed to execute quantization operations. This approach ensures that quantization regions remain distinct and do not overlap.
\end{itemize}

Algorithm \ref{algorithm} offers a visual representation of the proposed steps for clarity. In the following sections, we elaborate on these steps in detail.
\subsection{First Phase: Separation of Important Network Layers and Channels}
Our proposed method introduces a criterion to distinguish vital layers from non-important ones during quantization. This process enables the quantization of a DNN using a mixed-precision quantization technique. Inspired by weight pruning techniques \cite{https://doi.org/10.48550/arxiv.2003.03033}, our approach utilizes higher weight tensor values to identify important layers.
In essence, our analysis reveals that weights close to zero within each layer have a minimal impact on the accuracy of a DNN. Following a similar logic to weight pruning, we eliminate near-zero weights with minimal accuracy trade-offs. This strategy, guided by Eq. \ref{eq:layers}, designates layers with higher values as important layers, facilitating a substantial reduction in DNN size with marginal accuracy trade-offs.

Layers are partitioned based on the formula:
\begin{equation}
   F_l = \mathlarger \Sigma^{n}_{i=1}|W_{c_i}|
\label{eq:layers}
\end{equation}
where $W_{c_i}$ denotes the $i^{th}$ weight of the $c^{th}$ channel, and $n$ represents the number of convolution layers in a DNN. The absolute value in Eq. \ref{eq:layers} is used to emphasize the importance of the count of the layer weight without considering its sign. The $F_l$ function is computed for all layers of a network.

An alpha variable is defined to distinguish between important and non-important layers. This variable is user-defined and plays a crucial role in the resulting impact on the quantized network's accuracy and size. This user-specified alpha value serves to distinguish between important and non-important layers in the quantization process. Users have the flexibility to adjust the alpha value based on their specific needs or application requirements. The value chosen for alpha influences what percentage of layers should be considered important during quantization. This adaptability allows users to tailor the quantization process to prioritize either the accuracy or the size of the resulting quantized DNN, depending on their application context. To illustrate the effects of varying alpha values, Tables \ref{tab:alpha_mobilenetv2} and \ref{tab:alpha_resnet50} present the trade-offs between accuracy and size in the quantized DNN for different alpha values.
% \ref{tab:alpha_mobilenetv2}.

Once we've identified the important layers, the next step involves isolating the crucial channels from the non-essential ones. To achieve this, we employ Eq. \ref{eq:channels} to categorize individual channels:
% \begin{equation}
%     F_c = \sqrt{\mathlarger \Sigma^{n}_{i=1}{{(W_{c_i})}^2}}
% \label{eq:channels}
% \end{equation}
\begin{equation}
    F_c = \sqrt{\mathlarger W_{1}^2 + W_{2}^2 + ... + W_{n}^2}
\label{eq:channels}
\end{equation}
where $W_{n}$ represents the weights of channels within a layer. For instance, $W_{1}$ means the weights of first channel in a layer.
% and includes all of the channels over a layer.   

The $F_c$ function is computed for all channels within each layer and subsequently sorted in descending order. Similar to the logic behind layer selection, the channels of each layer are distinguished between important channels and others. A Beta variable is used to classify channels into two categories. If the selected layer is among the important layers, a larger proportion of channels from that layer is classified as important channels. Conversely, if the layer is non-important, most of its channels are deemed non-essential. This channel segregation allows for distinct bit precision quantization, enhancing quantization flexibility.

\RestyleAlgo{ruled}
%% This is needed if you want to add comments in
%% your algorithm with \Comment
\SetKwComment{Comment}{// }{}
\SetKwInput{KwInput}{Input}
\SetKwInput{KwOutput}{Output}
\SetKwFunction{FSort}{Sort}
\SetKwFunction{FQuantize}{Quantize}
\DontPrintSemicolon
\newcommand\mycommfont[1]{\footnotesize\ttfamily\textcolor{blue}{#1}}
\SetCommentSty{mycommfont}

\begin{algorithm}[hbt!]

\caption{Selection of important layers and channels with PTILMPQ Algorithm.}\label{algorithm}

\KwInput{Channel weights $x$ of a pre-trained model, $\alpha, \beta$ .}
\KwOutput{Separated layers/channels.}
$ImportantLayers, ImportantChannels,$

$ChannelList =$ [ ]

\ForEach{$l_i \in [L_1, ..., L_n]$}{\Comment{$l_i$ is a $i^{th}$ layer in a DNN.}
    $F_l = |W_{c_1}| + ... + |W_{c_n}|$ \Comment{Defined in Eq.\ref{eq:layers}}
    
    \If{$F_l > \alpha $}
        {$ImportantLayers \gets l_i$}
    \ForEach{$W_{c_i} \in [C_1, ..., C_n]$}{\Comment{$W_{c_i}$ is $i^{th}$  weight tensor in $c^{th}$ channel}
        $F_c = Sqrt({W_{1}}^2 + ... +{W_{2}}^2)$ 
        
        \Comment{ defined in Eq.\ref{eq:channels}}
        $ChannelList \gets F_{c}$
        
        Sort ($ChannelList$) \Comment{Ascending sort by weights}
        \eIf{$l_i \in ImportantLayers$}{
            $k = \beta \times length (ChannelList)$
            \Comment{$k$ is a boundary}
            $ImportantChannels \gets ChannelList[:k]$
        }{
            $k = (1 - \beta) \times length (ChannelList)$
            $ImportantChannels \gets ChannelList[:k]$
        }
    }
}

\end{algorithm}

\subsection{Second phase: Quantization using non-overlapping regions}
Following the initial phase, we distinguish important channels from their non-important counterparts, allocating higher bit precision to the list of vital channels. Consequently, the quantization bits are established during this first phase. 
In the second phase, we aim to minimize the potential accuracy loss in the neural network. To achieve this, we divide the quantization range for weight tensors within each neural network layer into two distinct, non-overlapping regions. 
In conventional quantization methods, the same weight range, determined by the maximum and minimum weights within each layer, is applied uniformly to all weight tensors, guiding the quantization process. However, in the proposed quantization method, which utilizes non-overlapping areas, we treat each weight tensor independently, designating two separate regions for each one. The quantization process is then executed based on these distinct regions, enhancing the precision of the quantization operation.

Consider a DNN characterized by an input weight tensor $x$ with floating-point precision and channel weights $W_c$. The quantization process entails the conversion of this neural network into a representation with $b$ bits. The quantization operation is executed as:
\begin{equation}
    x_q(x; m_x, M_x, b, z) = round \left( \frac{clamp(x; m_x, M_x) - z}{s} \right)
    \label{eq:main_quant}
\end{equation}
where $x$ is weight tensor, $x_q$ is the quantized weight with integer precision, $m_x$ and $M_x$ are minimum and maximum weight tensors respectively, $z$ is offset, $b$ is the number of quantization levels and $clamp()$ function and $s$ are defined:
\begin{equation}
    clamp(x; m_x, m_x) = min(max(x, m_x), M_x),
\end{equation}
\begin{equation*}
    s = \frac{M_x - m_x}{2^b - 1}.
\end{equation*}

\begin{figure}
 	    \vspace{-0.1in}
    \subfloat{\includegraphics[width=.5\columnwidth]{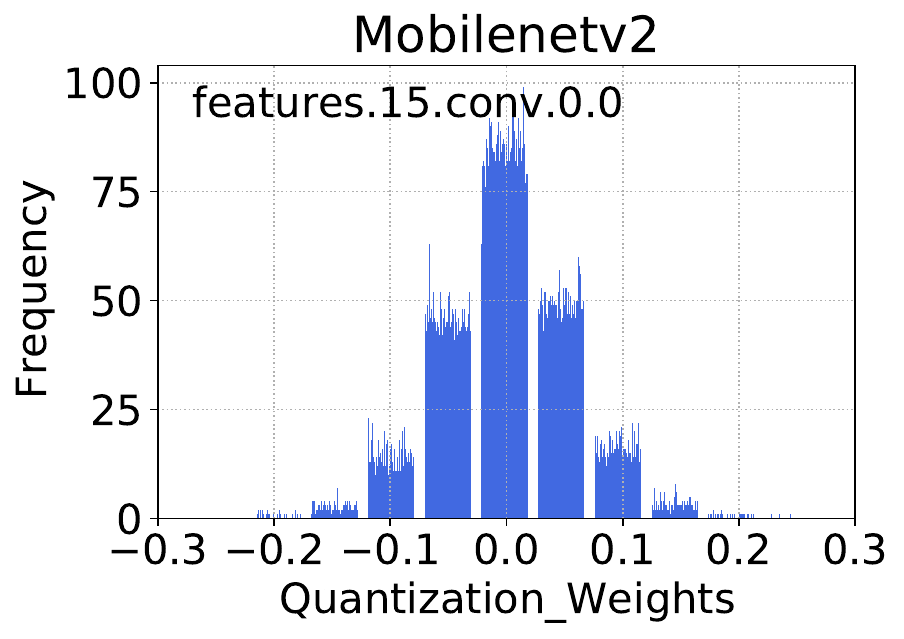}} 
    \subfloat{\includegraphics[width=.5\columnwidth]{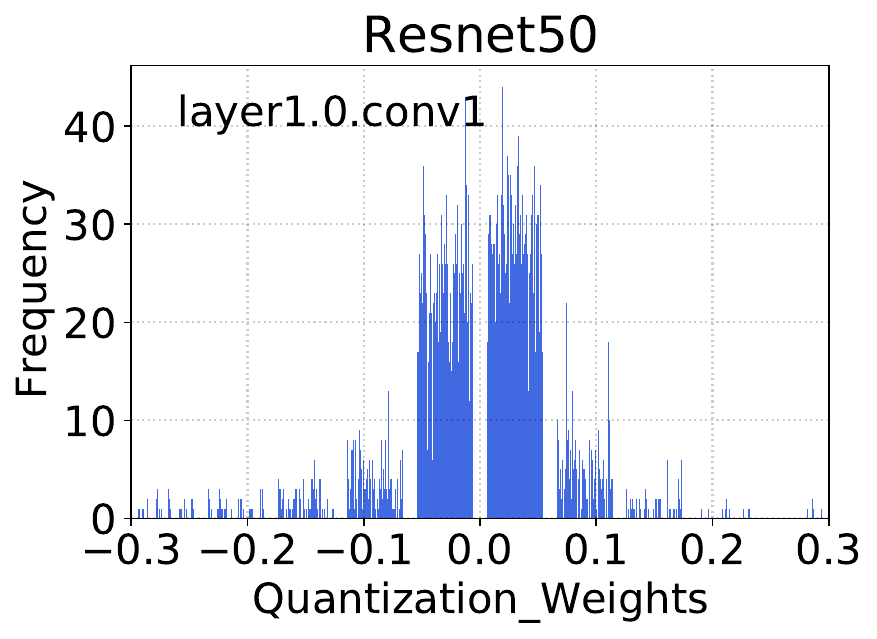}}
    \caption{Weight distribution for MobilenetV2 and Resnet50. As can be seen, they look bell-shaped.}
     	    \vspace{-0.1in}
    \label{fig:distribution}
\end{figure}

\begin{figure}
 	    \vspace{-0.02in}

    \centerline{\includegraphics[width=\columnwidth]{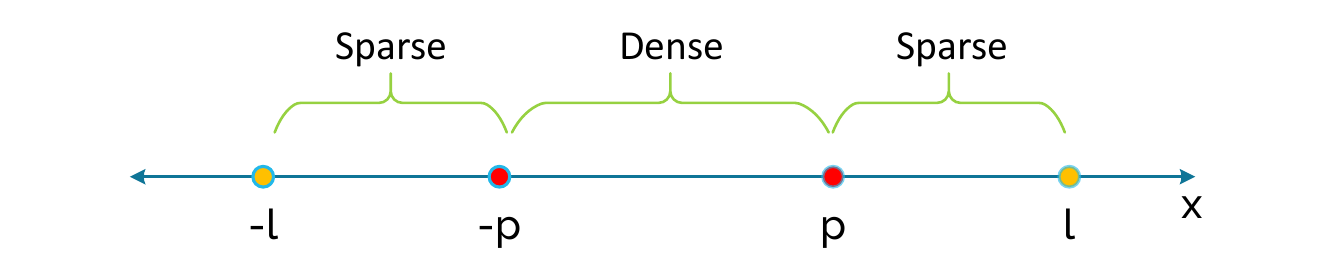}}
     	    \vspace{-0.1in}

    \caption{Dense-Region and Sparse-Region}
     	    \vspace{-0.1in}

    \label{fig:regions}
     	    \vspace{-0.0in}
\end{figure}

Note that in most DNNs, weight distributions typically exhibit a bell-shaped pattern (Figure \ref{fig:distribution} demonstrates this pattern for ResNet50 and MobileNetV2). Leveraging this observation, we divide the weight distribution of each layer into two non-overlapping intervals, forming the foundation for the quantization operation. This division is illustrated in Figure \ref{fig:regions}. In this figure the horizontal axis shows the size of the weight tensor of a layer and the vertical axis shows the number of repetitions of that weight. As can be seen in the figure, the data density in the central points is more than other areas. Using this point, the weight distribution interval for each layer is divided into two areas and the quantization operation is performed based on them. These two areas do not overlap with each other.

% It's worth noting that the weight distribution in the majority of DNNs exhibits a bell-shaped pattern.
% the weight distribution for two NNs ResNet50 and MobileNetv2 is shown in Fig.\ref{fig:distribution}.

Figure \ref{fig:regions} illustrates the division of these areas, with points \(p\) and \(-p\) marking the breaking points. The overall range of weight distribution is thereby divided into dense and sparse regions.

% Where points p and -p are breaking points and the overall range of weight distribution is divided into dense and sparse areas.

Determining the breaking point necessitates the calculation of quantization error. This quantization error arises from the mapping of floating-point values to integers and is computed as follows:
\begin{equation}
    e = x_{q} - x ,
\end{equation}
Therefore, the quantization error squared is defined as:
\begin{equation}
    E(e^2;b,m_{x},M_{x}) = \frac{s^2}{12} = C(b)\Delta^2,
\label{eq:main_error}
\end{equation}
\begin{equation*}
    \Delta = M_x - m_x
\end{equation*}
where, due to the uniformity of the quantization regions, the $C()$ function is defined:
\begin{equation}
C(b) = \frac{1}{12(2^b - 1)^2}
    \label{eq:cb}
\end{equation}
Using the new areas that were performed on the distribution of NN weights, quantization error squared is rewritten as:
\begin{equation}
    E(e^2;b,l,p)=C(b-1){(l-p)^2+l(2p-l)[2F(p)-1]}
\label{eq:new_error}
\end{equation}
where the function $F()$ is the cumulative distribution function. 

Using Eq.\ref{eq:new_error}, the breaking point is defined:
\begin{equation}
p^* =argmin_{ p \in (0,\frac{l}{2})} E(e^2;b,l,p).
\label{eq:breakpoint}
\end{equation}
The obtained breaking point is used for quantization operation and Eq.\ref{eq:main_quant} is rewritten:
\begin{equation}
    x_{q} (x; l, p, b, z) = \begin{cases} 
    sign(x) \times x_q(|x|; 0, p, b-1, 0), x \in D \\
    sign(x) \times x_q(|x|; p, l, b-1, p), x \in S
    \end{cases}
\label{eq:new_quant}
\end{equation}
where $D$ is Dense-Region and in the range $[-p, p]$, $S$ is Sparse-Region and in the range $[-l, -p) \cup (p, l]$, $l$ is upper bound $(l > 0)$, and $p \in [-l, l]$.
% (based on Fig.\ref{fig:regions})

In the conventional quantization process, all network weights are uniformly converted into their respective integers using Eq. \ref{eq:main_quant}. However, in non-overlapping quantization, the weights within a single layer are categorized into two distinct regions based on their values: the sparse region and the dense region. It's important to highlight that, despite this division into regions, the quantization operation is carried out with identical quantization boundaries for both regions, classifying the proposed method under the umbrella of uniform quantization.

% In the normal quantization operation, all the NN weights are converted to the corresponding integer using Eq.\ref{eq:main_quant}. While in non-overlapping quantization for the weights of one layer, two different regions are considered according to their value (sparse and dense region). Note that the quantization operation is performed for each region with the same quantization region, and this indicates that the proposed method is in the category of uniform quantization.

\section{Experimental Results}
\label{sec:results}

In this section, we present the results of the proposed quantization experiments on two widely adopted models, ResNet50 and MobileNetV2. The quantization process is applied to the ImageNet dataset, employing precision levels of 8-bit, 6-bit, 4-bit, 3-bit, and 2-bit using our proposed method. Throughout all our experiments, we maintain a consistent 8-bit precision for activation functions. Additionally, we employ batch normalization fold techniques \cite{jacob2018quantization} to enhance DNN accuracy before quantization. Pre-trained models are utilized with the PyTorchVision 0.13.1 framework. Subsequently, we illustrate the trade-off between accuracy and memory size in the following section. Finally, we delve into benchmarking BOPs across varying precisions to provide a comprehensive assessment.

% In this section, the results of the quantization operation on two popular models, ResNet50 and MobileNetV2, are analyzed. The quantization operation is performed on the ImageNet dataset with 8-bit, 6-bit, 4-bit, 3-bit and even 2-bit precision using the proposed method. In all of our experiments, 8-bit precision has been used for activation functions. Also, batch normalization fold \cite{jacob2018quantization} has been used to increase the accuracy of the NNs before performing the quantization operation. To use pre-trained models, Pytorchvision 0.13.1 has been used. In the next section, the trade-off between the accuracy and the size of the NNs is displayed in the form of a graph. At the end, the benchmark of BOPs with different precision is discussed.

\subsection{Evaluating ResNet50 and MobileNet Testcases}
Tables \ref{tab:resnetnew} and \ref{tab:mobilenetnew} present the outcomes of our proposed quantization method alongside other approaches, focusing on accuracy and the size of the neural network. We employ a segment of the ImageNet training dataset, referred to as the calibration dataset, for initializing the activation functions. Notably, our proposed method obviates the need for a retraining step after quantization, making it independent of a training dataset. The ImageNet test dataset serves as the testing benchmark for our method.

In Table \ref{tab:resnetnew}, "MP" denotes the use of mixed precision quantization, wherein 8-bit precision is applied to important layers and 2-bit precision to non-important layers. Additionally, we utilize 6-bit and 4-bit precision for important layers and 2-bit precision for non-important layers.
The results demonstrate that our proposed quantization method allows for the quantization of non-important layers with reduced precision, effectively diminishing the size of the neural network while maintaining accuracy. In the case of 8-bit precision quantization mode, our method achieves a 37.74\% reduction in neural network size compared to the PWLQ method, which employs a fixed number of quantization bits but relies on non-overlapping regions. This size reduction can be attributed to our approach's ability to allocate different bit precisions based on the identification of important layers and channels using Eq. \ref{eq:layers} and Eq. \ref{eq:channels}. Utilizing varying bit precisions significantly impacts the size of the neural network, as non-important layers and channels are quantized with fewer bits, ultimately reducing the model size. It's important to note that the ZEROQ method considers quantization bit precisions for weights as {2, 4, 8}; however, for assessing the number of neural network parameters post-quantization, a 4-bit quantization precision is assumed.

\begin{table*}[tb!]
    \centering
     	    \vspace{-0.1in}

        \setlength{\extrarowheight}{2pt}
        \caption{
        %Uniform quantization, PWLQ\cite{fang2020post}, QWP\cite{krishnamoorthi2018quantizing}, SSBD\cite{meller2019same}.
        Our method and other methods with Resnet50 on Imagenet with 8-bit quantization for activation and mixed precision for weights. (8,2)-MP means using 8-bit precision for important channels and 2-bit precision for non-important channels.}
    \begin{tabular}{c c c c c c}
    \Xhline{1pt}
        method & precision (W/A) & Top-1 & Top-5 & Size(Mbit) & Size Reduction (\%)\\ 
        \hline\hline
        Baseline & - & 76.13 & 92.86 & 816 & - \\
        \hline\hline
        Uniform Q & 8 / 8 & 76.06 & 92.88 & 204 & 75.00\\
        PWLQ \cite{fang2020post} & 8 / 8 & 76.10 & 92.86 & 204 & 75.00 \\
        QWP\cite{krishnamoorthi2018quantizing} & 8 / 8 & 75.10 & - & 204 & 75.00 \\
        SSBD \cite{meller2019same} & 8 / 8 & 74.95 & - & 204 & 75.00 \\
        \rowcolor{mycolor}
        % \hline
        Ours & (8,2)-MP / 8 & 75.65 & 92.73 & \textbf{127} & \textcolor{red}{\textbf{84.43}} \\ [1ex] 
        \hline\hline
        Uniform Q & 6 / 8 & 75.89 & 92.87 & 153 & 81.25 \\
        PWLQ\cite{fang2020post} & 6 / 8 & 76.08 & 92.90 & 153 & 81.25 \\
        \rowcolor{mycolor}
        % \hline
        Ours & (6,2)-MP / 8 & 74.40 & 92.25 & \textbf{114} & \textcolor{red}{\textbf{86.02}} \\ [1ex]
        \hline\hline
        Uniform Q & 4 / 8 & 72.75 & 91.09 & 102 & 87.50 \\ 
        PWLQ \cite{fang2020post} & 4 / 8 & 75.62 & 92.72 & 102 & 87.50\\
        QWP \cite{krishnamoorthi2018quantizing} & 4 / 8 & 54.00 & - & 102 & 87.50 \\
        ZEROQ \cite{cai2020zeroq} & MP / 8 & 75.80 & - & 97.48 & 88.05\\
        \rowcolor{mycolor}
        % \hline
        Ours & (4,2)-MP / 8 & 74.41 & 92.22 & \textbf{88} & \textcolor{red}{\textbf{89.21}}\\ [1ex]
        
    \Xhline{1pt}
    \end{tabular}
    \label{tab:resnetnew}
\end{table*}

To counteract the substantial accuracy dip observed in the MobileNetV2 model when subjected to 2-bit precision quantization, we opted for a more robust 3-bit precision approach to safeguard the DNN's accuracy. Table \ref{tab:mobilenetnew} provides insights into the quantization processes executed on the MobileNetV2 model, utilizing 8-bit, 6-bit, and 4-bit precision for important layers, and 3-bit precision for non-important layers. When juxtaposed with the non-quantized model in 8-bit mode, our proposed quantization method reveals a remarkable 81.23\% reduction in the DNN model's size, with only a slight accuracy decrease of less than 1\%. This outcome underscores the ability of our approach to significantly shrink the size of the MobileNetV2 while preserving its accuracy, despite its already compact architecture.

\begin{table*}[tb!]
    \centering
        \setlength{\extrarowheight}{2pt}
        \caption{{Comparison of the proposed method of quantization with other methods based on accuracy and size of MobileNetV2.}}

    \begin{tabular}{c c c c c c }
    \Xhline{1pt}
        method & precision (W/A) & Top-1 & Top-5 & Size(Mbit) & Size Reduction(\%) \\ 
        \hline\hline
        
        Baseline & - & 71.87 & 90.28 & 111.03 & - \\ \hline\hline
        Uniform Q & 8 / 8 & 71.55 & 90.19 & 27.73 & 75.00\\
        PWLQ \cite{fang2020post} & 8 / 8 & 71.73 & 90.23 & 27.73 & 75.00 \\
        QWP \cite{krishnamoorthi2018quantizing} & 8 / 8 & 69.80 & - & 27.73 & 75.00 \\
        SSBD \cite{meller2019same} & 8 / 8 & 71.19 & - & 27.73 & 75.00 \\
        \rowcolor{mycolor}
        % \hline
        Ours & (8,3)-MP / 8 & 70.78 & 89.71 & \textbf{20.84} & \textcolor{red}{\textbf{81.23}} \\[1ex]
        \hline\hline
        Uniform Q & 6 / 8 & 70.85 & 89.82 & 20.81 & 81.25\\
        PWLQ \cite{fang2020post} & 6 / 8 & 71.75 & 90.25 & 20.81 & 81.25 \\
        % \hline
        \rowcolor{mycolor}
        Ours & (6,3)-MP / 8 & 70.70 & 89.69 & \textbf{16.67} & \textcolor{red}{\textbf{84.98}}\\ [1ex]
        \hline\hline
        PWLQ \cite{fang2020post} & 4 / 8 & 69.22 & 88.43 & 13.87 & 87.50 \\
        QWP \cite{krishnamoorthi2018quantizing} & 4 / 8 & 71.80 & - & 13.87 & 87.50 \\
        ZEROQ \cite{cai2020zeroq} & MP / 8 & 68.83 & - & 13.36 & 87.96 \\
        DDQ \cite{zhang2021differentiable} & 4 / 8-MP & 71.90 & - & 14.40 & 87.03 \\
        \rowcolor{mycolor}
        Ours & (4,3)-MP / 8 & 67.50 & 87.90 & \textbf{11.72} & \textcolor{red}{\textbf{89.44}} \\ [1ex]
        
        \Xhline{1pt}
    \end{tabular}
     	    \vspace{-0.0in}

    \label{tab:mobilenetnew}
\end{table*}

\subsection{Impact of the Alpha Parameter}
In the context of single precision mode, modifying the bit precision for quantization operations proves impractical. However, PTILMPQ introduces a notable degree of flexibility, allowing the quantization bit precision to be finely tuned according to the specific demands of the given application. This method essentially empowers users to make judicious decisions regarding quantization, taking into account factors such as DNN size, constrained by memory limitations imposed by available resources. Furthermore, in instances where the emphasis is on accuracy or DNN size reduction, users have the liberty to select the quantized model with the desired precision. The manipulation of alpha values, as elucidated in Algorithm \ref{algorithm}, serves as a key mechanism for effectively fine-tuning the trade-off between DNN size and accuracy. By adjusting alpha, users can assign a higher value for scenarios prioritizing accuracy and a lower value when the focus is on reducing DNN size. Consequently, the quantization operation dynamically adjusts to the specified alpha value. For a comprehensive understanding of the impact of alpha adjustments on accuracy and DNN size, refer to Tables \ref{tab:alpha_resnet50} and \ref{tab:alpha_mobilenetv2} for the ResNet50 and MobileNetV2 models, respectively.

% In single precision mode, it is not possible to change the bit precision for quantization operations. But the proposed method has a great flexibility and the quantization bit precision can be changed based on the desired application. In other words, the proposed method makes it possible to choose the appropriate quantization in situations where the NN size is important (memory limitation) based on available memory. In addition, for the case where accuracy has a higher priority than the size of the NN, the quantized model with high accuracy should be selected. By using alpha changes (Algorithm \ref{algorithm}), it is possible to change the priority between the size and accuracy of the NN. Therefore, by using alpha changes, if accuracy is important, the value of alpha can be considered high, and if the size of the NN has a higher priority than the accuracy of the NN, a low value alpha is considered and then the quantization operation is performed based on the alpha value. Tables \ref{tab:alpha_resnet50} and \ref{tab:alpha_mobilenetv2} show the changes of alpha and its effect on the accuracy and size of the NN for the ResNet50 and MobileNetV2 models, respectively. 

\begin{table}[tb!]
    \centering
        \setlength{\extrarowheight}{2pt}
        \caption{Quantization results of MobileNetV2 on ImageNet. A trade-off is seen between the accuracy and size of the NN when considering different $\alpha$ (layer selection criteria).}
    \begin{tabular}{||c c c c||}
    \hline
        precision & $\alpha$ & Top-1(\%) & size(Mbit) \\ [0.5ex]
        \hline\hline
        \rowcolor{mycolor}
        W4 & - & 68.53 & 13.87\\
        \hline
        W4 MP & 5 & 63.22 & \textcolor{red}{\textbf{11.22}} \\
        W4 MP & 10 & 63.34 & 11.22\\
        W4 MP & 15 & 64.15 & 11.24 \\
        W4 MP & 20 & 65.26 & 11.25 \\
        W4 MP & 25 & 66.99 & 11.55 \\
        W4 MP & 30 & \textcolor{red}{\textbf{67.28}} & 11.57 \\
        \hline
    \end{tabular}
    \label{tab:alpha_mobilenetv2}
     	    \vspace{-0.0in}
\end{table}

\begin{figure*}[tb!]
    \centering
    \vspace{-0.1in}

    \subfloat[Precision 2-bit and 8-bit]{\includegraphics[width=0.32\textwidth]{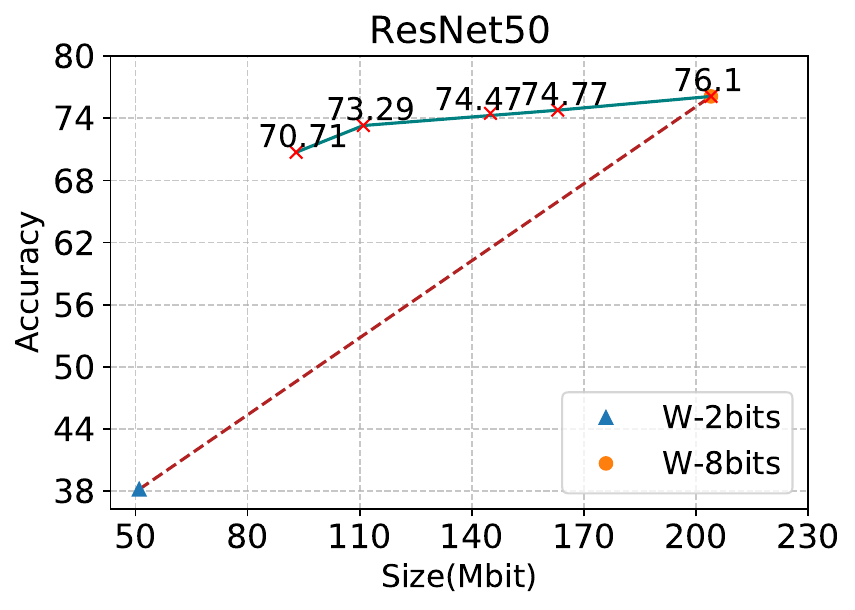}}
    \hfill
    \subfloat[Precision 2-bit and 6-bit]{\includegraphics[width=0.32\textwidth]{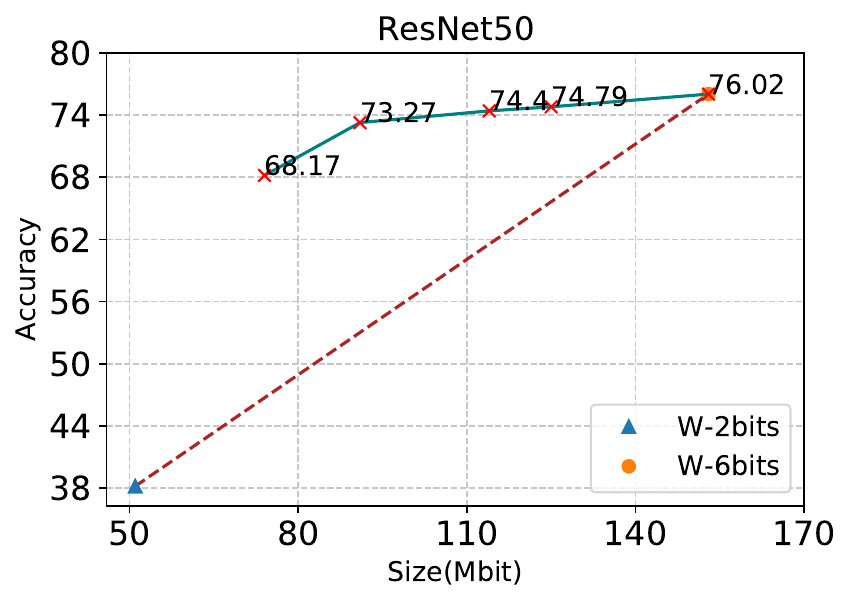}}
    \hfill
    \subfloat[Precision 2-bit and 4-bit]{\includegraphics[width=0.32\textwidth]{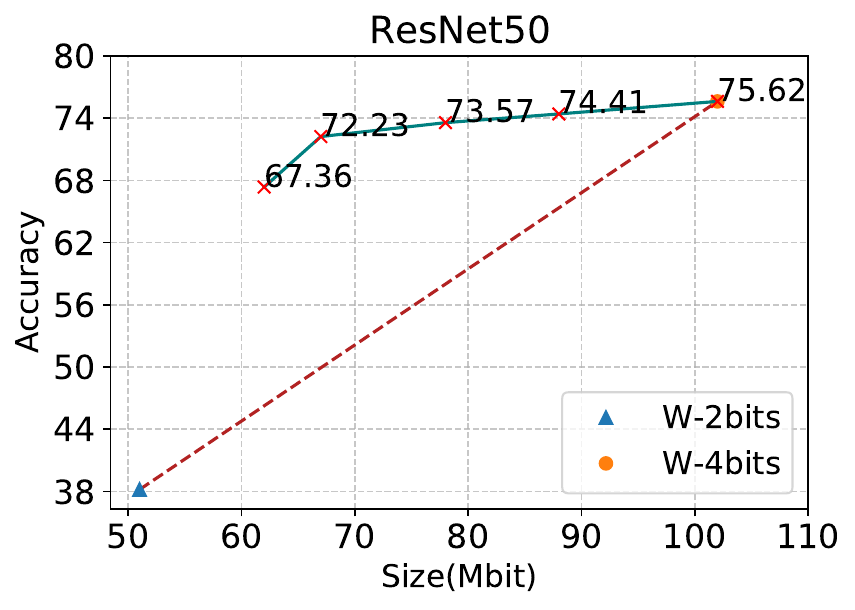}}

    \vspace{0.1in} % Adjust the vertical space between rows

    \subfloat[Precision 3-bit and 8-bit]{\includegraphics[width=0.32\textwidth]{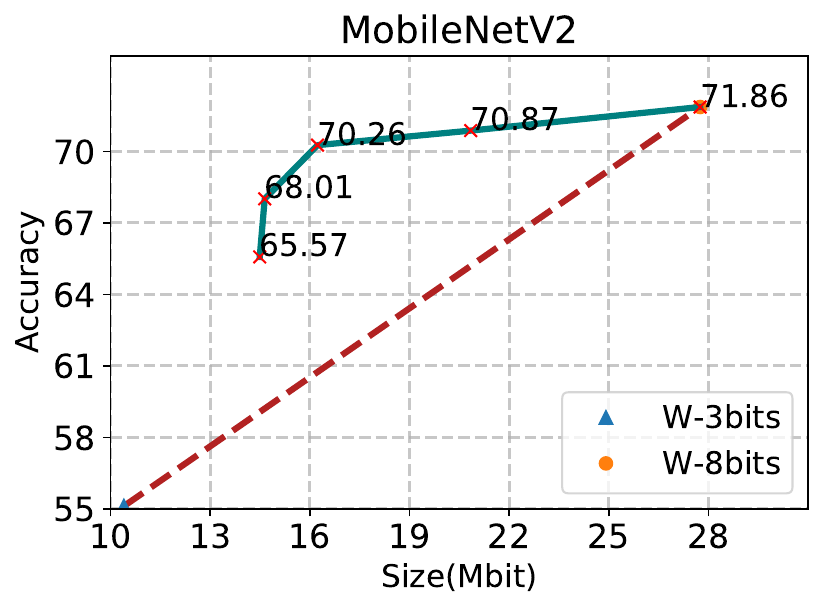}}
    \hfill
    \subfloat[Precision 3-bit and 6-bit]{\includegraphics[width=0.32\textwidth]{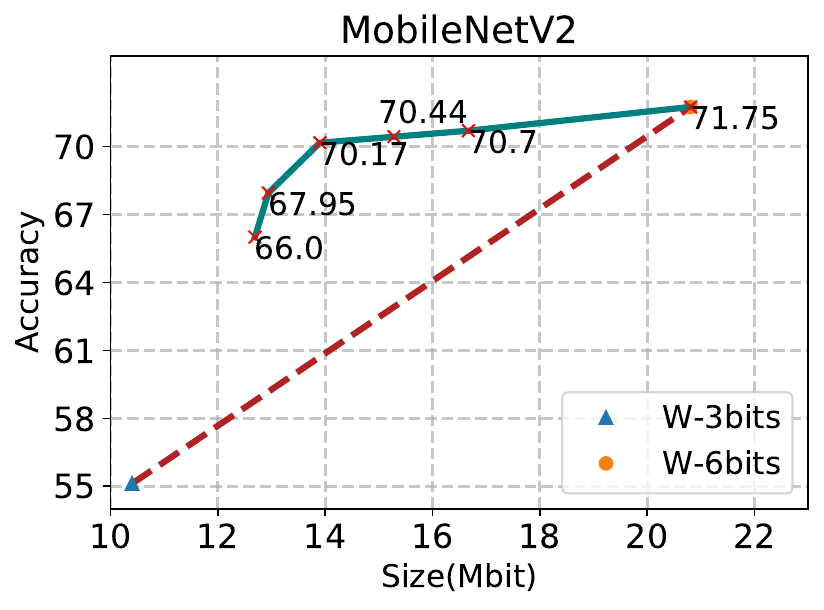}}
    \hfill
    \subfloat[Precision 3-bit and 4-bit]{\includegraphics[width=0.32\textwidth]{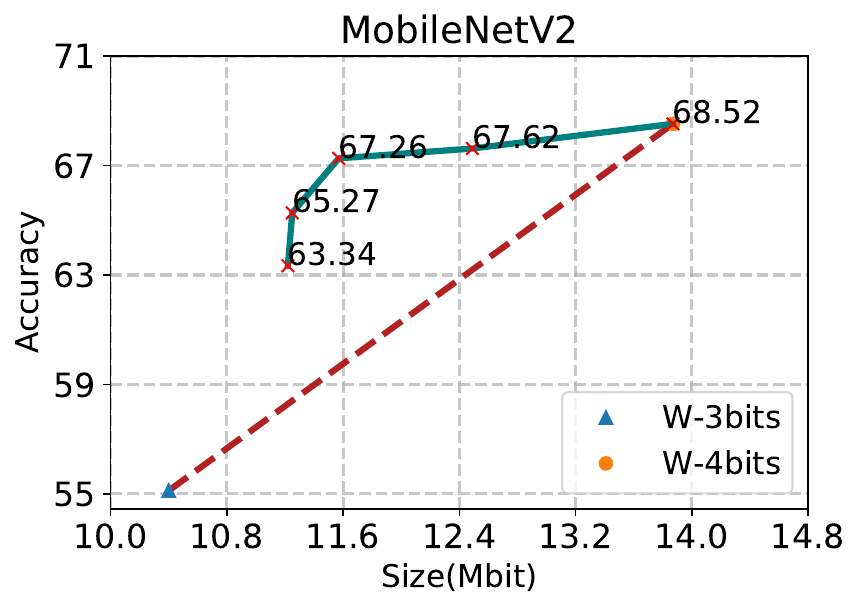}}

    \caption{Trade-off between accuracy and model size. Dotted line represents quantization with single-precision, and the blue line represents mixed-precision quantization with different alphas.}
    \label{fig:tradeoff_size_acc}
    \vspace{-0.1in}
\end{figure*}

Figure \ref{fig:tradeoff_size_acc} reveals the trade-off between NN size and accuracy for ResNet50 and MobileNetV2 models. The x-axis indicates NN size in Mbit, while the y-axis represents Top1 accuracy. The dotted line signifies fixed bit precision quantization results in NN size and precision. The intersection points represent quantization outcomes using our flexible method, allowing precision and size adaptations via alpha variations. Rows a to c display ResNet50 quantization results for different alpha values, comparing them to fixed 2-bit and 8-bit precision quantization. Rows d to f demonstrate MobileNetV2 quantization outcomes with various alpha values, showcasing our method's adaptability through alpha adjustments.

% Figure \ref{fig:tradeoff_size_acc} shows the trade-off between NN accuracy and NN size on ResNet50 and MobileNetV2 models.
% The horizontal and vertical axes show the size of the NN based on Mbit and the Top1 accuracy of the NN, respectively. The beginning and end of the dotted line are the precision and size of the NN using quantization operations with fixed bit precision.  The crossing points are the quantization of the NN using the proposed method, in which the precision and size of the NN are changed by using alpha change. In Figure \ref{fig:tradeoff_size_acc}, the first row (from a to c) shows the quantization operation based on the proposed method on the ResNet50 model, which Figure 6.a shows the comparison of quantization with different alpha values between two quantization with fixed bit precision of 2-bit and 8-bit. The second row (from d to f) shows the quantization operation based on the proposed method with different alpha values on MobileNetV2 NN. In this row, the flexibility of the proposed method is shown based on alpha changes. 

% \begin{figure*}[tb!]
%  	    \vspace{-0.1in}

% \centering
%     \subfloat[Precision 2-bit and 8-bit]{\includegraphics[width=0.50\columnwidth]{Figs2/tradeoff-resnet50-82v5.pdf}}
%     \subfloat[Precision 2-bit and 6-bit]{\includegraphics[width=0.50\columnwidth]{Figs2/tradeoff-resnet50-62v2.pdf}}
%     \subfloat[Precision 2-bit and 4-bit]{\includegraphics[width=0.50\columnwidth]{Figs2/tradeoff-resnet50-42v8.pdf}}
    
%     \subfloat[Precision 3-bit and 8-bit]{\includegraphics[width=0.50\columnwidth]{Figs2/tradeoff-mobilenetv2-83v3.pdf}}
%     \subfloat[Precision 3-bit and 6-bit]{\includegraphics[width=0.50\columnwidth]{Figs2/tradeoff-mobilenetv2-63v6.pdf}}
%     \subfloat[Precision 3-bit and 4-bit]{\includegraphics[width=0.50\columnwidth]{Figs2/tradeoff-mobilenetv2-43v3.pdf}}
%     \caption{Trade-off between accuracy  and model size. Dotted line is quantization with single-precision and blue line is mixed-precision quantization with different alphas.}
%     \label{fig:tradeoff_size_acc}
%      \vspace{-0.1in}
% \end{figure*}

\subsection{Assessment of Computational Complexity}
\label{sub:Bops}
In the realm of quantized neural networks, the conventional metric of FLOPs (Floating-Point Operations) becomes obsolete, given the replacement of floating-point numbers with integer precision. Instead, we adopt BOPs (Bit Operations) as an alternative measure, offering a more relevant and insightful assessment of computational complexity by considering the number of bits involved. The BOPs formula is given by \cite{baskin2021uniq}:
\begin{equation}
    BOPs = mnk^2 (b_a b_w + b_a + b_w \times log_2 nk^2)
    \label{eq:bops}
\end{equation}
where $b_a$ and $b_w$ are the bit precision used for activation and weight respectively, $n$, $m$ and $k$ are number of input channel, number of output channel and kernel size respectively.

\begin{figure*}[h!]
    \centering
    \subfloat{\includegraphics[width=0.4\textwidth]{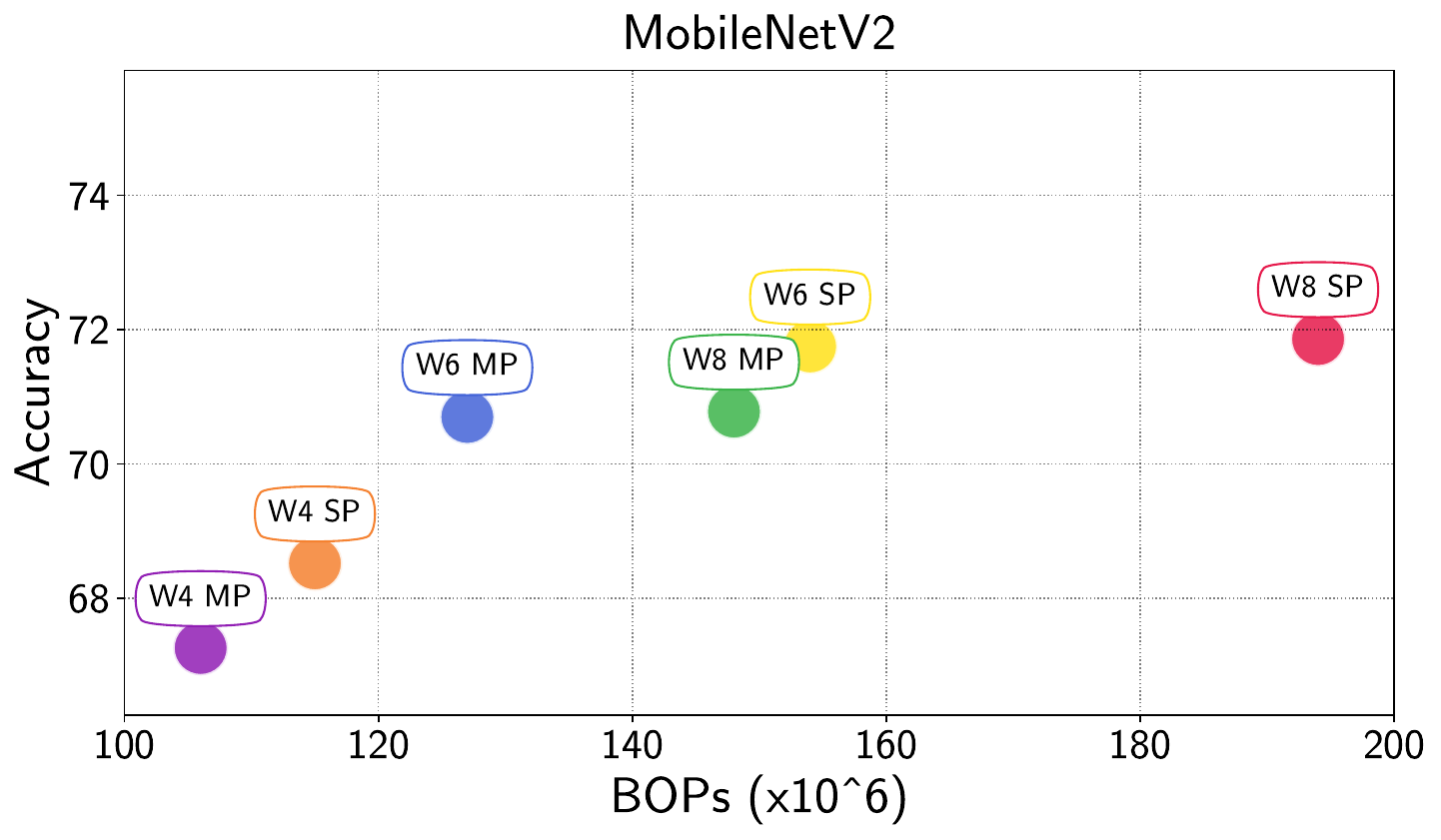}} 
    \subfloat{\includegraphics[width=0.4\textwidth]{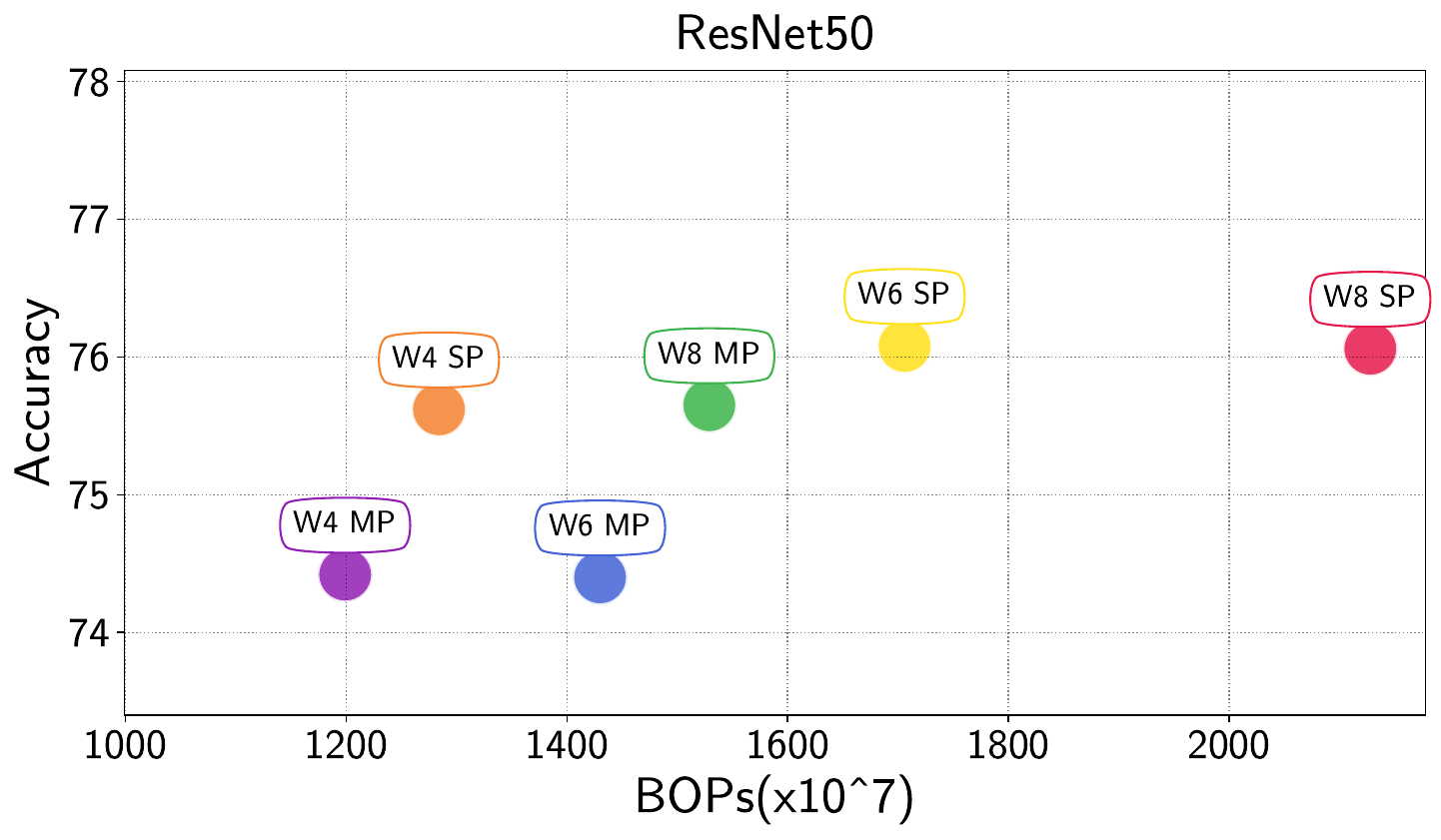}}
     \vspace{-0.1in}
    \caption{The graph showcases the computed BOPs for both Single-Precision (SP) and our proposed Mixed-Precision (MP) method, providing insight into the trade-off between accuracy and computational complexity for MobileNetV2 and ResNet50 architectures.}
    \label{fig:bops}
     	    \vspace{-0.2in}
\end{figure*}

It is paramount to assess the performance of variable bit quantization in quantized neural networks. Figure \ref{fig:bops} provides a visual representation of BOPs for the proposed method across a spectrum of bit precisions. This comparison extends to fixed-precision approaches, encompassing both ResNet50 and MobileNetV2 models. In this figure, "SP" designates Single Precision, while "MP" represents Mixed Precision. In mixed precision mode, crucial layers operate with the specified bit precision, while less critical layers adopt 2-bit precision for ResNet50 and 3-bit precision for MobileNetV2. For example, the notation "W8 MP" in ResNet50 signifies the use of 8 bits for important layers and 2 bits for non-important layers during the quantization process. This visual representation serves as a comprehensive overview of the computational intricacies associated with different bit precisions in our approach.

\begin{table}[h!]
    \centering
        \setlength{\extrarowheight}{2pt}
        \caption{Quantization results of ResNet50 on ImageNet.
        % As described in Section \ref{sec:method},
        $\alpha$ is a criterion for selecting important layers from unimportant ones. There is a trade-off between accuracy and model size.}

    \begin{tabular}{||c c c c||}
    \hline
        precision & $\alpha$ & Top-1(\%) & size(Mbit) \\ [0.5ex]
        \hline\hline
        \rowcolor{mycolor}
        W4 & - & 75.69 & 102\\\hline
        W4 MP & 5 & 67.36 & \textcolor{red}{\textbf{62.88}} \\
        W4 MP & 15 & 72.20 & 66.96 \\
        W4 MP & 10 & 70.10 & 65.12 \\
        W4 MP & 20 & 72.84 & 71.20 \\
        W4 MP & 30 & 73.92 & 82.56 \\
        W4 MP & 40 & \textcolor{red}{\textbf{74.41}} & 88.40 \\\hline
    \end{tabular}
     	    \vspace{-0.1in}
    \label{tab:alpha_resnet50}
\end{table}
\section{Conclusion}
\label{sec:concl}

This paper tackles the growing demand for deploying DNN models on resource-constrained edge devices. We have introduced a novel technique, Post-Training Intra-Layer Multi-Precision Quantization (PTILMPQ), meticulously crafted to minimize DNN memory footprints, making them well-suited for deployment on edge devices. PTILMPQ adopts a post-training quantization approach, eliminating the requirement for extensive training data, and incorporates precise bit allocation by estimating the importance of layers and channels within the network.
The quantitative results underscore the efficiency of the proposed method in achieving notable reductions in model size across different neural network architectures. The utilization of mixed precision, coupled with the nuanced consideration of layer importance and weight distribution characteristics, positions PTILMPQ as a promising approach for practical deployment, especially in resource-constrained environments where minimizing model size is a critical factor.

% In this paper, we introduced a new post training quantization method for deep neural networks. Two approaches are used to estimate important layers and channels and to categorize them into important and non-important categories. After this, we quantize the important category with high-precision while the unimportant category is quantized with low-precision. As part of the quantization process we use the non-overlapping area technique to improve the accuracy of the model. 

% The final step was to test the quantized model on an Android phone and evaluate it by measuring metrics such as accuracy and memory consumption during the program using the TensorFlow Lite framework.

\bibliographystyle{IEEEtran}
\bibliography{references}

% \bibliography{references}	
% \end{thebibliography}

% \bibliographystyle{ACM-Reference-Format}

\end{document}